\documentclass[runningheads]{llncs}

\usepackage{accv}

\usepackage{accvabbrv}

\usepackage{graphicx}
\usepackage{booktabs}
\usepackage{amsmath,amssymb}
\usepackage{multirow}
\usepackage{xcolor}
\usepackage{colortbl}
\usepackage{enumitem}
\usepackage{placeins}  
\usepackage[accsupp]{axessibility}

\usepackage[breaklinks,colorlinks,citecolor=accvblue]{hyperref}
\usepackage{orcidlink}

\definecolor{oursgray}{RGB}{236,240,245}
\newcommand{\best}[1]{\textbf{#1}}
\newcommand{\secbest}[1]{\underline{#1}}

\newcommand{\blfootnote}[1]{%
  \begingroup
  \renewcommand\thefootnote{}\footnote{#1}%
  \addtocounter{footnote}{-1}%
  \endgroup
}

\begin{document}

\title{Learning Features, Not Layouts: Feature-Axis Attention as a Transfer Prior for Medical Imaging}

\titlerunning{Feature-Axis Attention as a Transfer Prior for Medical Imaging}

\author{Aon Safdar\orcidlink{0000-0003-3039-1017} \and
Mohamed Saadeldin\orcidlink{0000-0003-4104-9524}}
\authorrunning{A. Safdar and M. Saadeldin}
\institute{University College Dublin, Ireland\\
\email{aon.safdar@ucdconnect.ie, mohamed.saadeldin@ucd.ie}}

\maketitle

\begin{abstract}
Natural-image-pretrained foundation models are increasingly adapted to medical imaging, yet transfer remains unreliable under domain shift, modality heterogeneity, and label scarcity. We trace part of this fragility to an overlooked design choice. Standard vision transformers place all global modeling on self-attention over \emph{spatial tokens}, an object-layout prior well suited to natural scenes but not to medical images, where the discriminative signal lies in feature correlations and texture that recur across scanners while layout does not. We ask whether the \emph{axis} of global attention, spatial versus feature, is itself a lever for transferable medical representations. Existing channel-aware hybrids confound this question, adding channel attention to a spatially-mixed backbone and leaving the axes coupled and the feature axis unmeasurable in isolation. We design FAD-ViT, a backbone that strictly decouples the two axes, making feature-axis attention the primary global operator and the attention axis a controlled variable. Evaluating comprehensively across 23 classification, segmentation, and low-data tasks under matched tokenization, capacity, and protocol, we find that shifting global attention from the spatial to the feature axis leaves from-scratch performance essentially unchanged, yet markedly strengthens transfer. Identical ImageNet pretraining yields up to roughly three times the transfer gain, with consistent improvements on every held-out segmentation task at up to a quarter the parameter count and far lower compute than standard and channel-attention baselines. The advantage is largest where texture rather than geometry is discriminative, tracking greater reuse of pretrained features. The axis of global attention is thus a design lever for medical transfer, and making it channel-primary is a simple, cheap, and principled choice.
\end{abstract}

\section{Introduction}
\label{sec:intro}
\blfootnote{Code and configurations are available at \url{https://github.com/aonsafdar/FAD-ViT}.}

Foundation-model pipelines increasingly dominate medical image analysis: a strong encoder pretrained on large-scale natural images is adapted to downstream clinical tasks~\cite{qu_adapting_2025,khan_recent_2025}. Despite advances in parameter-efficient adaptation and decoder design~\cite{dutta_parameter_nodate,ruffini_benchmarking_2026}, transfer remains inconsistent under domain shift, acquisition heterogeneity, limited supervision, and dense prediction~\cite{ali_task-specific_2026}. Most transferred encoders inherit the standard vision transformer (ViT)~\cite{dosovitskiy_vit_2021}, in which global modeling is dominated by self-attention over \emph{spatial tokens} while feature/channel interaction is left to a mostly static feed-forward network. This bias suits natural-image pretraining, where object shape and layout are discriminative, but is less appropriate for medical transfer, where pathology often manifests as subtle feature correlations rather than dominant shape cues and exact spatial layout varies across devices, protocols, and anatomy.

The root of this mismatch is that spatial self-attention is asked to do two distinct jobs with one operator: \emph{extract} local spatial structure and \emph{mix} information globally. It is well matched to neither, carrying only a weak locality prior and costing \(\mathcal{O}(N^2)\) in the token count. Hybrid and hierarchical backbones~\cite{wu_cvt_2021,liu_swin_2021,wang_pyramid_2021} add locality and efficiency but keep spatial attention as the primary global operator, and even channel-aware models that make the \emph{feature} axis dynamic~\cite{elnouby_xcit_2021,ding_davit_2022} graft it onto a pipeline still built for spatial mixing, changing only the mixing axis while inheriting spatial tokenization, projections, and locality modules. A different design separates the two jobs outright: extract with cheap depthwise convolution and mix globally with attention over \emph{channel} tokens, so that each token summarizes the whole image and global interaction becomes an input-dependent reweighting of features at linear spatial cost. The consequences of this channel-primary design for \emph{medical transfer} are unstudied, and are the subject of this paper: whether making global interaction channel-primary changes how well a backbone carries pretrained structure into heterogeneous medical domains, and if so, why.

\begin{figure}[tb]
\centering
\includegraphics[width=1\textwidth]{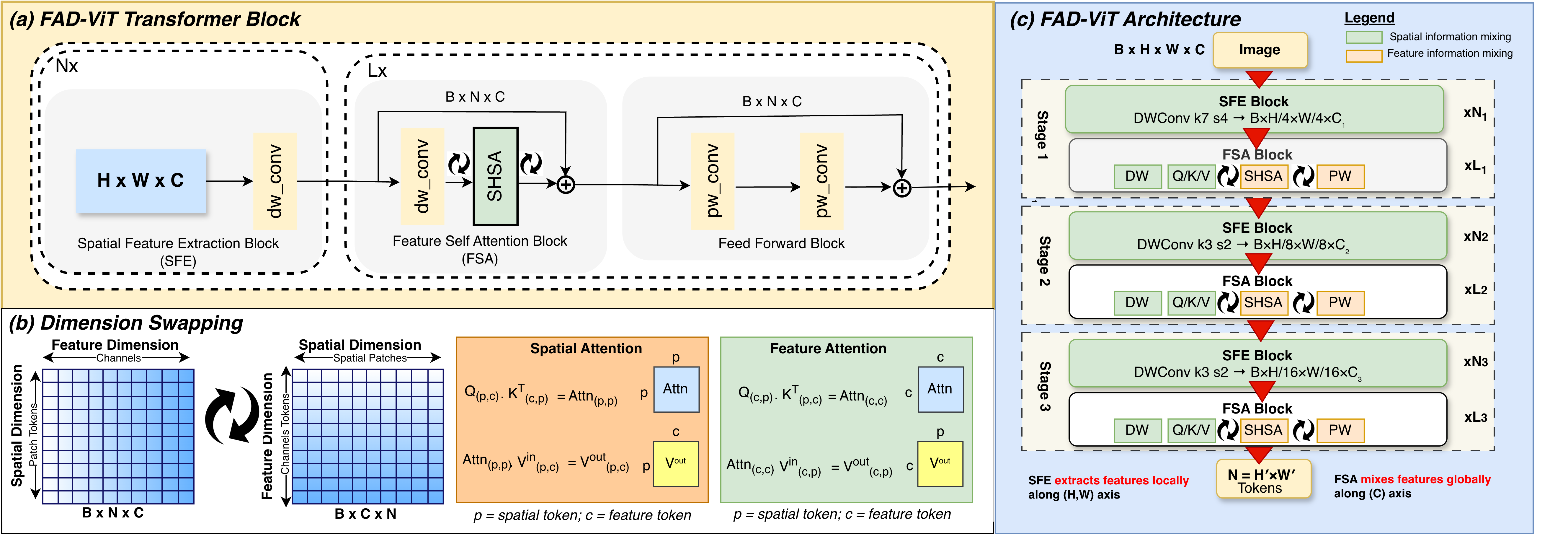}
\caption{Overview of the studied block. Spatial Feature Extraction (SFE) preserves the 2D grid with depthwise convolution; Feature Self-Attention (FSA) transposes tokens (\(B{\times}N{\times}C \rightarrow B{\times}C{\times}N\)) and attends over feature tokens. Global interaction thus scales with width \(C\) rather than token count \(N\).}
\label{fig:main}
\end{figure}

\noindent\textbf{What FAD-ViT is.}
FAD-ViT is not a spatial backbone with channel attention added as a branch; it is a strictly axis-decoupled, \emph{channel-primary} backbone that separates the two conflated jobs onto purpose-matched operators. Depthwise convolution extracts local structure and keeps the 2D grid intact through depth, while single-head self-attention over \emph{channel} tokens, each a feature-map descriptor summarizing the whole image, performs global mixing at \(\mathcal{O}(C^2 N)\) cost, linear in the token count \(N\), rather than the \(\mathcal{O}(N^2C)\) of spatial attention. Because even a pointwise (\(1{\times}1\)) projection mixes channels, \(Q/K/V\) are formed by \emph{depthwise-only} projections, leaving channels unmixed before the attention that correlates them; this strict separation, which prior channel-aware models forgo (\cref{sec:related}), is what lets us treat the global axis as an isolable variable. Channel-primary mixing is already global at linear spatial cost, so the design needs no windowing, local-interaction, dual-axis, or gating module, and defers a short spatial-attention tail to the lowest resolution where it is cheap.

\noindent\textbf{Contribution and scope.}
Our contribution is a controlled account of \emph{why} a strictly decoupled, channel-primary design is the right prior for medical transfer, and a characterization of how it behaves across modalities, tasks, and label budgets. Concretely:
\begin{itemize}[nosep,leftmargin=1.2em]
\item We show that strict axis decoupling with channel-primary mixing acts as a \emph{transfer prior} for medical imaging: it converts natural-image pretraining into substantially larger downstream gains than a spatial-axis backbone of identical capacity, because pretraining on the channel axis deposits inter-feature correlations (texture, intensity, co-activation) that survive medical domain shift, whereas spatial-layout priors do not (\cref{sec:analysis}).
\item We characterize FAD-ViT across 23 medical tasks, classification, segmentation, and low-data probing, under one matched protocol, and against the coupled channel designs XCiT and DaViT, showing that the decoupled design transfers competitively or better at a fraction of their parameters (\cref{sec:results,sec:decouple}).
\item We isolate the axis with matched from-scratch versus pretrained comparisons and representation-reuse analyses, well-posed only \emph{because} the axes are cleanly separated here and not entangled as in prior channel-aware backbones (\cref{sec:transfer-spec,sec:why}).
\end{itemize}

\section{Related Work and Positioning}
\label{sec:related}

\noindent\textbf{Hierarchical and hybrid vision transformers.}
Staged backbones such as CvT~\cite{wu_cvt_2021}, Swin~\cite{liu_swin_2021}, and PVT~\cite{wang_pyramid_2021} introduce convolutional tokenization, windowed attention, and pyramidal downsampling for locality and efficiency. These designs retain \emph{spatial-token} attention as the dominant global operator. FAD-ViT shares their hierarchical, convolution-tokenized structure but is built around channel-primary mixing rather than spatial attention (\cref{sec:method}). For the controlled study, this shared skeleton lets us also swap only the global-attention axis while holding tokenization, capacity, and schedule fixed, attributing transfer differences to the axis rather than to pyramid structure.

\noindent\textbf{Channel-axis and dual-axis attention.}
Because attention is a content-adaptive mixing operator, it can mix channels rather than spatial tokens. Four design dimensions of a transformer stage are useful to separate: (i) tokenization, (ii) non-attention projections (locality modules and the MLP), (iii) attention projections (\(Q/K/V\) formation), and (iv) the mixing axis. XCiT~\cite{elnouby_xcit_2021} introduces cross-covariance attention over channels, so its \emph{primary} mixer is already the feature axis, yet it changes only dimension (iv), retaining single-scale tokenization (i), a depthwise local-patch-interaction module with pointwise MLP (ii), pointwise \(Q/K/V\) (iii), and a terminal spatial class-attention stage. DaViT~\cite{ding_davit_2022} keeps spatial attention co-equal, alternating windowed spatial and grouped channel attention with pointwise \(Q/K/V\). Later channel-aware backbones follow the same template, interleaving, fusing, or gating the two axes rather than separating them: SCAT~\cite{zheng_lightweight_2023}, DCT-ViT~\cite{guan_dynamic_2025}, SCHEME~\cite{sridhar_scheme_2024}, and asymmetric factorized bilinear operators~\cite{wu_asymmetric_2025} all enrich channel mixing inside an otherwise spatial pipeline. Since dimensions (i)--(iii) are inherited from spatial-attention baselines, spatial and feature interaction stay entangled: pointwise \(Q/K/V\) pre-mix channels before the feature attention, and spatial mixing recurs throughout. FAD-ViT instead co-designs all four dimensions around channel mixing: staged convolutional tokenization (i), depthwise spatial extraction with \emph{no} channel mixing plus a pointwise MLP (ii), depthwise-\emph{only} \(Q/K/V\) (iii), unlike the pointwise projections above and the depthwise-separable projections of CvT~\cite{wu_cvt_2021} that add redundant pre-attention channel mixing, and channel-primary global mixing with spatial attention confined to a short low-resolution tail (iv). It therefore needs none of the windowing~\cite{liu_swin_2021}, local-interaction~\cite{elnouby_xcit_2021}, dual-axis~\cite{ding_davit_2022}, or gating machinery these models rely on, and is the only design in which no operator both extracts space and mixes channels. To our knowledge no prior hybrid or channel-aware backbone strictly decouples the two axes in this way: the channel-mixing projections or residual spatial attention they retain keep the axes entangled. Strict decoupling enables making the feature axis attributable; it lets us isolate feature-axis (channel) attention as the \emph{sole} global operator under matched tokenization, capacity, and protocol, and characterize it as a transfer prior for medical imaging, an attribution the entangled designs above cannot support.

\noindent\textbf{Transformers for medical imaging.}
Medical transformers commonly retain spatial attention as the primary global operator and add channel or dual-branch modules~\cite{linguraru_med-former_2024,zhang_dual-attention_2024,wang_feat_nodate}, sparsify attention~\cite{linguraru_revisiting_2024}, or fuse convolutional and transformer features~\cite{chen_medfusenet_2025,li_cfformer_2026} as secondary refinements; others swap spatial attention for cheaper global mixers~\cite{dang_logvmamba_2024,xie_msumlp_2024}. We invert this hierarchy, making feature-axis attention the primary global operator, and evaluate it as a transfer prior rather than a task-specific module. A large self-supervised encoder (DINOv2~\cite{oquab_dinov2_2024}) serves as a high-capacity baseline under the same protocol. We take XCiT and DaViT as the transfer comparators because they are the canonical representatives of the two channel-aware families, a pure channel-axis backbone and an interleaved dual-axis one, and provide public ImageNet weights; more recent channel-aware models refine these same two templates rather than decoupling the axes, so the comparison generalizes to them.

\section{Method}
\label{sec:method}

\subsection{Axis-Decoupled Block}
FAD-ViT uses a three-stage convolutional-tokenization hierarchy but, unlike spatial and dual-axis backbones, assigns spatial extraction and global mixing to two \emph{purpose-matched}, disjoint operators rather than folding both into spatial attention. At stage \(s\), let the feature map be \(\mathbf{X}^{(s)}\in\mathbb{R}^{B\times H_s\times W_s\times C_s}\), reshaped to tokens \(\mathbf{X}^{(s)}_{\mathrm{tok}}\in\mathbb{R}^{B\times N_s\times C_s}\) with \(N_s=H_sW_s\). Each block applies the two decoupled operations below; for the controlled study we also form a spatial-axis counterpart by swapping only the global operator while holding tokenization, capacity, and schedule fixed.

\noindent\textbf{Spatial Feature Extraction (SFE).}
A depthwise convolution refines local structure while preserving the spatial grid,
\begin{equation}
\mathbf{U}^{(s)} = \mathrm{DWConv}\big(\mathbf{X}^{(s)}\big),
\end{equation}
with stage-dependent kernel and stride. The convolution is depthwise \emph{by design, not for efficiency}: because it applies an independent kernel per channel, all spatial mixing is confined to this branch and no channels are combined here, so spatial modeling is cleanly separated from feature modeling. Local spatial structure is thus routed by SFE and only by SFE, which keeps spatial contiguity for dense prediction and avoids premature spatial--feature entanglement.

\noindent\textbf{Feature Self-Attention (FSA).}
Global interaction runs along the feature axis. Queries, keys, and values are produced by a \emph{depthwise} projection that acts on each channel independently, so channels are never linearly combined before attention; the resulting affinity is therefore computed between unmixed channel identities, in contrast to the pointwise (channel-mixing) query/key/value projections of cross-covariance attention~\cite{elnouby_xcit_2021}. The tensor is then flattened and transposed from token layout to feature layout,
\begin{equation}
B{\times}C_s{\times}H_s{\times}W_s \;\xrightarrow{\text{flatten}}\; B{\times}N_s{\times}C_s \;\xrightarrow{\text{transpose}}\; B{\times}C_s{\times}N_s ,
\end{equation}
so that the \(C_s\) channels become the tokens and the \(N_s\) spatial positions become the feature dimension. Each channel token is thus the flattened \(H_s{\times}W_s\) response map of a single channel, i.e.\ a complete view of the image, so the full 2D layout is carried \emph{inside} every token and is restored exactly when the output is transposed back; attention reweights whole-image channel responses and never collapses spatial structure. Single-head attention over this layout,
\begin{equation}
\mathrm{FSA}(\mathbf{Q},\mathbf{K},\mathbf{V})
= \mathrm{Softmax}\!\left(\frac{\mathbf{Q}\mathbf{K}^{\top}}{\sqrt{N_s}}\right)\mathbf{V},
\qquad
\mathbf{Q},\mathbf{K},\mathbf{V}\in\mathbb{R}^{B\times C_s\times N_s},
\label{eq:fsa}
\end{equation}
produces a \(C_s{\times}C_s\) feature-affinity map, in contrast to the \(N_s{\times}N_s\) token-affinity map of spatial attention. The map recombines the per-channel value maps, so each output token is a data-dependent reweighting of whole-image channel responses rather than of spatial locations. The result is transposed back to \(B{\times}N_s{\times}C_s\) and updated with residual connection and feed-forward mixing. Because a feature token already aggregates the full spatial field, partitioning it across attention subspaces buys little and a single head suffices, as in the channel-token single-head attention of DaViT~\cite{ding_davit_2022} and the single-head macro design of SHViT~\cite{yun_shvit_2024}; unlike DaViT's grouped variant we keep one global head over the full channel set.

Together SFE and FSA partition the two forms of interaction into disjoint operators: SFE mixes space but not channels, FSA mixes channels but does not collapse space, and no operator in the block does both. This clean separation is what makes the global axis a variable we can set and study in isolation.

\subsection{Staged Backbone, Spatial Tail, and Complexity}
The backbone stacks three stages with progressive downsampling and channel expansion \((H_1,W_1,C_1)\!\to\!(H_2,W_2,C_2)\!\to\!(H_3,W_3,C_3)\), \(C_1{<}C_2{<}C_3\). The final stage appends an optional, lightweight \emph{spatial-attention tail} (a few multi-head spatial-token blocks) on top of the feature-axis backbone. This tail is a minor refinement rather than a co-equal mixer; \cref{sec:analysis} shows that removing it costs only \(\approx\)1 Dice, so feature-axis attention performs the overwhelming majority of global mixing and the backbone remains channel-primary by design.

The two axes have complementary cost. Spatial attention scales as \(\mathcal{O}(N_s^2 C_s)\), quadratic in the token count \(N_s=H_sW_s\) and therefore in input resolution; feature attention scales as \(\mathcal{O}(C_s^2 N_s)\), quadratic in width \(C_s\) but only linear in \(N_s\). The feature-axis cost per position is thus fixed by architecture and independent of resolution, whereas the spatial-axis cost grows quadratically once tokens outnumber channels (\(N_s>C_s\)). At the moderate token counts of this backbone (\(N_s\!\approx\!C_s\) at deep stages) the two are comparable in FLOPs; the feature axis's decisive advantage here is parametric: FAD-ViT reaches its transfer accuracy with \(23.1\)M parameters against \(90.5\text{--}90.6\)M for the ViT/DINOv2 decoders (\cref{sec:results}), while its resolution-independent global cost avoids the spatial axis's blow-up at the high native resolutions common in medical acquisition. For segmentation we attach an identical U-Net decoder to the staged features across all encoders, so that any downstream difference is attributable to the encoder rather than the decoder.

\section{Experimental Setup}
\label{sec:setup}

\begin{figure}[tb]
\centering
\includegraphics[width=0.84\textwidth]{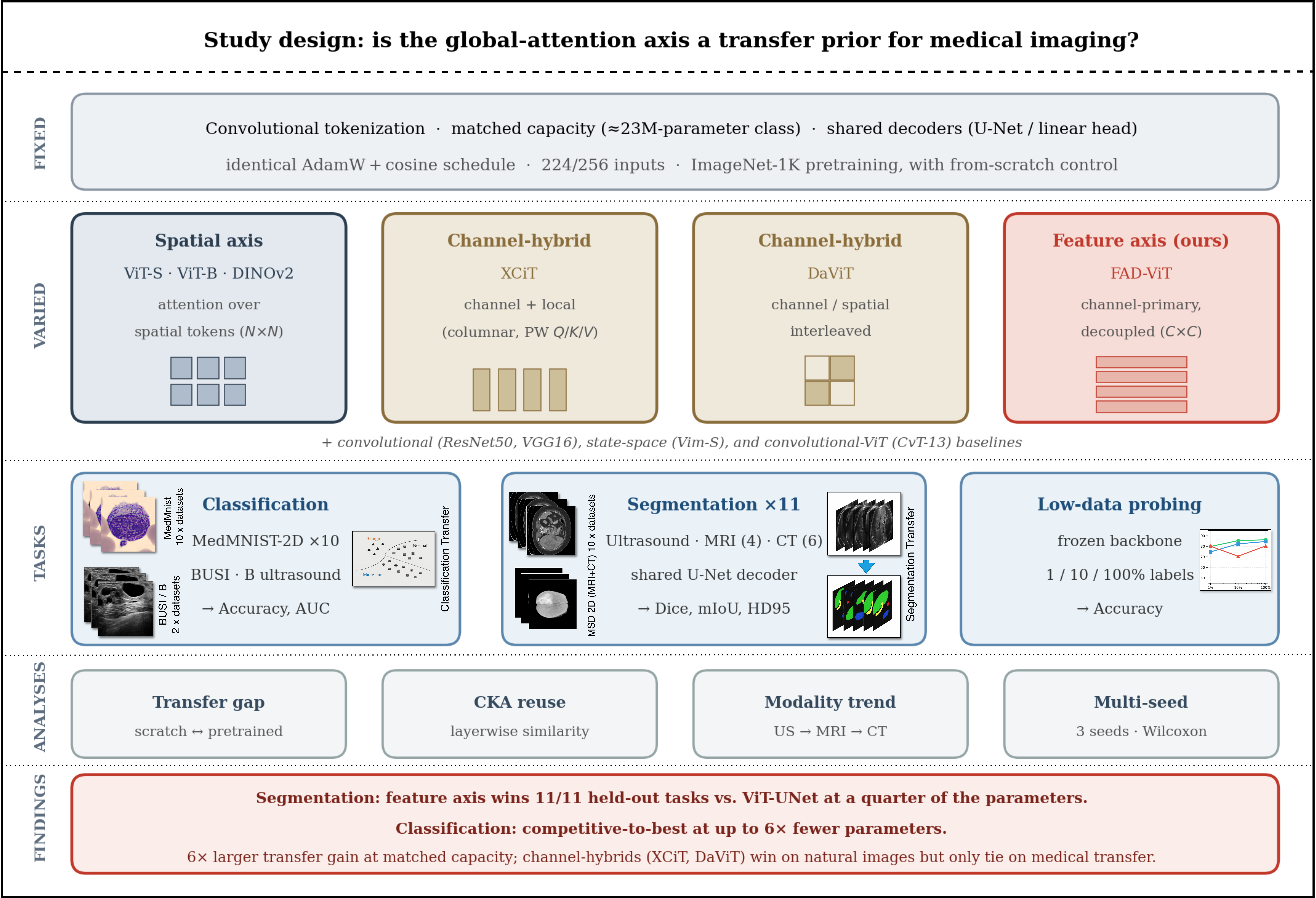}
\caption{Study design. Under a fixed protocol (convolutional tokenization, matched \(\sim\)23M-parameter capacity, shared U-Net / linear-head decoders, identical schedule, and ImageNet-1K pretraining with a from-scratch control) we vary \emph{only} the global-mixing operator across four backbone families: a spatial-axis baseline (ViT/DINOv2), two channel-hybrids (XCiT, DaViT), and the strictly decoupled feature axis. Transfer is evaluated on classification, segmentation, and low-data probing and attributed with transfer-gap, CKA, modality, and multi-seed analyses.}
\label{fig:protocol}
\end{figure}

\noindent\textbf{Protocol.}
To isolate encoder representation quality we fix everything downstream: identical ImageNet-1K pretraining (except DINOv2, LVD-142M~\cite{oquab_dinov2_2024}), identical optimizer, schedule, augmentation, input resolution, and, for segmentation, an identical U-Net decoder across encoders (\cref{fig:protocol}). This controls for pipeline effects so that differences reflect the backbone.

\noindent\textbf{Datasets.}
We evaluate on 23 medical datasets/tasks: (i) \emph{classification} on 10 MedMNIST-2D datasets~\cite{yang_medmnist_2023} plus BUSI~\cite{al-dhabyani_dataset_2020} and dataset~B~\cite{yap_automated_2018} breast ultrasound; (ii) \emph{segmentation} on BUSI lesion and 10 MSD-derived 2D tasks~\cite{antonelli_medical_2022}; (iii) \emph{low-data linear probing} on 8 MedMNIST datasets at 1\%, 10\%, and 100\% labels. For the MSD-derived 2D tasks the $70/15/15$ split is made at the volume (patient) level \emph{before} slicing, so all slices of a volume stay in one partition and no patient spans splits; BUSI is split at the image--mask-pair level.

\noindent\textbf{Baselines.}
We compare against spatial-attention ViTs (ViT-S/16, ViT-B), convolutional ViT (CvT-13~\cite{wu_cvt_2021}), a convolutional U-Net encoder (ResNet50), other CNNs (VGG16), a state-space model (Vim-S), and a large self-supervised encoder (DINOv2-B/14~\cite{oquab_dinov2_2024}). For the axis study specifically we compare feature-axis (ours) against a matched-capacity spatial ViT (ViT-S/16-UNet, \(22.9\)M) and against the canonical channel-axis and dual-axis designs XCiT~\cite{elnouby_xcit_2021} and DaViT~\cite{ding_davit_2022} within the same protocol.

\noindent\textbf{Implementation.}
Classification models are fine-tuned end-to-end; segmentation encoders share a U-Net decoder and are fine-tuned; linear probing freezes the backbone and trains a linear head. Training uses AdamW with cosine scheduling and mixed precision; input is \(224{\times}224\) for classification and \(256{\times}256\) for segmentation. We report Accuracy/AUC (classification), Dice/mIoU/HD95 (segmentation), and linear-probe accuracy at each label fraction. Full preprocessing per modality (intensity windowing, resampling, normalization) and all hyperparameters are given in the supplementary material. All runs use fixed seeds and an H100/A100 GPU pool.

\section{Results}
\label{sec:results}

\subsection{Classification Transfer}
\Cref{tab:medmnist} reports MedMNIST-2D transfer. On these small, mostly saturated classification benchmarks the three ImageNet-pretrained encoders are close: FAD-ViT is best or second-best on all ten datasets, but ViT-S/16 leads on several; we do not read these narrow margins as the primary evidence, since the axis effect is clearest under dense prediction and the transfer-gap analysis (\cref{sec:transfer-spec}), where capacity is stressed and layout priors are put under domain shift. \Cref{tab:busi} reports breast-ultrasound transfer, a harder, texture-dominated target: FAD-ViT obtains the best accuracy on BUSI and B and best AUC on B. Notably it does so at or below the parameter count of the matched-capacity ViT-S/16 baseline, and matches or surpasses substantially larger encoders, including DINOv2-B/14 (evaluated at the same \(224\)px as every model), rather than the comparison resting on any single large baseline.

\begin{table}[tb]
\centering
\caption{MedMNIST-2D classification transfer (ImageNet-1K pretraining). Test Accuracy/AUC (\%). \best{Bold}: best; \secbest{underline}: second.}
\label{tab:medmnist}
\setlength{\tabcolsep}{4pt}\footnotesize
\begin{tabular}{lcccccc}
\toprule
\multirow{2}{*}{Dataset} & \multicolumn{2}{c}{CvT-13} & \multicolumn{2}{c}{ViT-S/16} & \multicolumn{2}{c}{FAD-ViT (feat-axis)}\\
\cmidrule(lr){2-3}\cmidrule(lr){4-5}\cmidrule(lr){6-7}
 & Acc & AUC & Acc & AUC & Acc & AUC\\
\midrule
PathMNIST   & 98.50 & \best{99.90} & \secbest{99.00} & \best{99.90} & \best{99.40} & \best{99.90}\\
DermaMNIST  & 73.77 & \secbest{91.32} & \secbest{76.50} & \best{93.00} & \best{77.20} & 90.70\\
OCTMNIST    & 91.36 & 97.27 & \best{94.39} & \best{98.42} & \secbest{93.44} & \secbest{98.30}\\
RetinaMNIST & 54.17 & 74.46 & \secbest{56.67} & \secbest{78.89} & \best{57.50} & \best{79.54}\\
BreastMNIST & 67.52 & 77.95 & \secbest{87.18} & \best{93.60} & \best{89.74} & \secbest{92.54}\\
BloodMNIST  & 93.68 & 99.46 & \best{96.57} & \best{99.83} & \secbest{96.45} & \secbest{99.68}\\
TissueMNIST & 67.04 & 92.66 & \secbest{72.19} & \secbest{94.21} & \best{73.47} & \best{94.60}\\
OrganAMNIST & 98.55 & 99.97 & \best{99.81} & \secbest{99.99} & \secbest{99.31} & \best{100.00}\\
OrganCMNIST & 95.11 & 99.76 & \best{98.64} & \best{99.96} & \secbest{98.33} & \secbest{99.85}\\
OrganSMNIST & 83.87 & 98.83 & \secbest{91.46} & \secbest{99.34} & \best{93.31} & \best{99.46}\\
\bottomrule
\end{tabular}
\end{table}

\begin{table}[tb]
\centering
\caption{Breast-ultrasound transfer on BUSI and B (shared fine-tuning). Params in millions; Accuracy/AUC (\%). \best{Bold}: best; \secbest{underline}: second.}
\label{tab:busi}
\setlength{\tabcolsep}{5pt}\footnotesize
\begin{tabular}{lccccc}
\toprule
Encoder & Params & \multicolumn{2}{c}{BUSI} & \multicolumn{2}{c}{B}\\
\cmidrule(lr){3-4}\cmidrule(lr){5-6}
 & (M) & Acc & AUC & Acc & AUC\\
\midrule
ResNet50    & 23.5  & 85.64 & 93.23 & 78.33 & 90.05\\
VGG16       & 134.3 & 85.47 & 93.69 & 80.00 & 88.24\\
ViT-S/16    & 21.7  & 86.50 & 93.60 & 82.50 & \secbest{90.68}\\
ViT-S/32    & 22.5  & 84.10 & 93.59 & 79.17 & 87.55\\
CvT-13      & 20.1  & 75.50 & 88.90 & 75.22 & 89.17\\
Vim-S/16    & 25.4  & \secbest{87.86} & \best{95.63} & \secbest{84.17} & 87.42\\
DINOv2-B/14 & 85.7  & 86.84 & \secbest{94.23} & 84.01 & 89.42\\
\rowcolor{oursgray}
FAD-ViT (feat-axis) & 21.7 & \best{88.21} & 93.75 & \best{87.50} & \best{92.57}\\
\bottomrule
\end{tabular}
\end{table}

\subsection{Segmentation Transfer}
\begin{figure}[tb]
\centering
\includegraphics[width=0.8\textwidth]{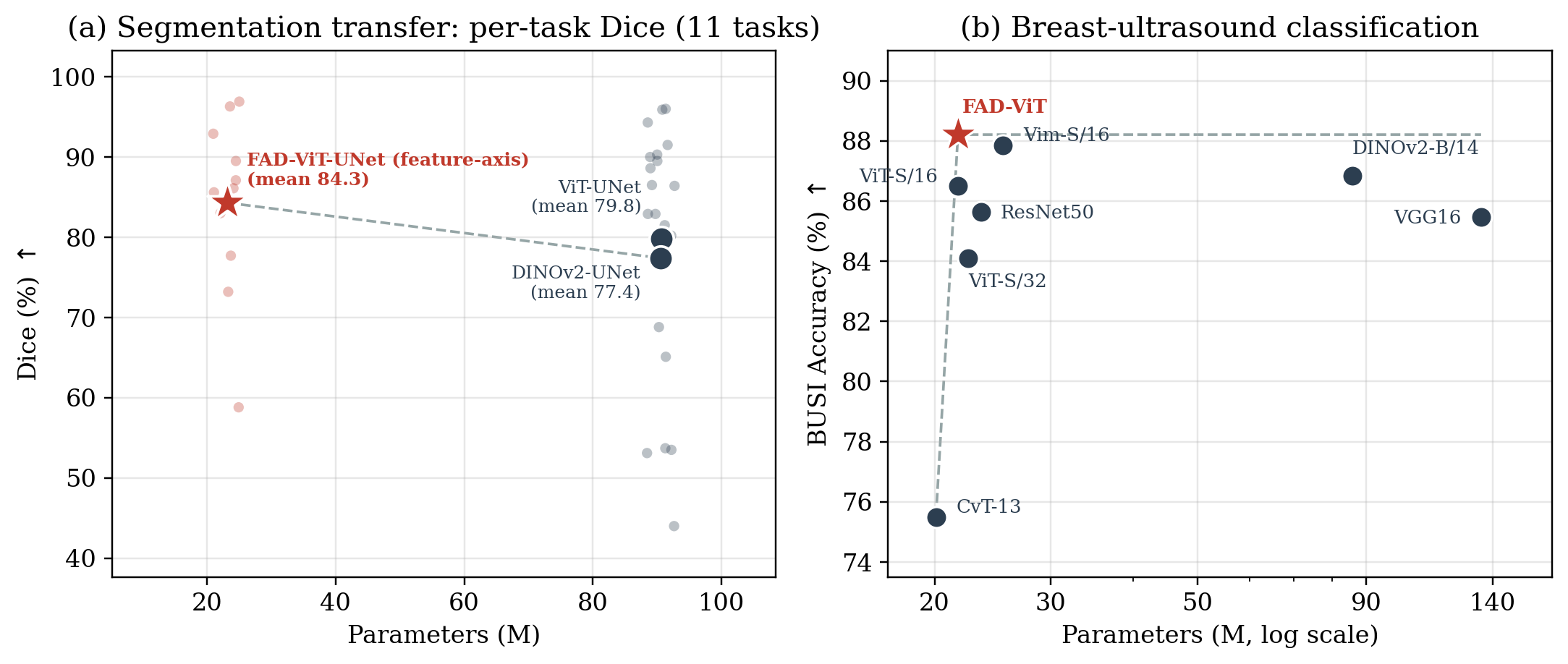}
\caption{Accuracy--efficiency trade-off under the matched transfer protocol. (a) Segmentation transfer: faint points are the 11 per-task Dice scores per encoder, large markers the mean; (b) BUSI classification accuracy versus parameters (log scale) across eight encoders spanning \(20\text{--}134\)M. Feature-axis FAD-ViT (\textcolor[RGB]{192,57,43}{$\star$}) is Pareto-dominant in both, matching or exceeding encoders up to \(6\times\) larger (DINOv2-B, VGG16; all evaluated at the same \(224\)px input) at a fraction of the parameters.}
\label{fig:pareto}
\end{figure}

\Cref{tab:seg} reports segmentation on the held-out test split (15\% of each dataset, disjoint from the training and validation partitions used for epoch selection), with the matched U-Net decoder. Feature-axis FAD-ViT-UNet attains the best Dice on ten of the eleven tasks and the best mean (84.3 versus 79.8 for ViT-UNet and 77.4 for DINOv2-UNet); only the near-saturated colon CT task marginally favors DINOv2. The advantage is systematic rather than driven by a few tasks: a one-sided Wilcoxon signed-rank test over the eleven paired tasks rejects equality against ViT-UNet (\(11/11\) wins, mean \(+4.4\) Dice, \(95\%\) bootstrap CI \([1.8,8.8]\), \(p\!=\!4.9\times10^{-4}\)) and against DINOv2-UNet (\(10/11\), \(+6.9\), \([3.3,11.3]\), \(p\!=\!1.5\times10^{-3}\)). Boundary quality improves in tandem (lower HD95 on most tasks), indicating fewer outlier errors rather than only better overlap. The largest gains fall on texture-dominant, low-contrast targets, headed by breast ultrasound (\(+24.6\) Dice over ViT-UNet, \(+20.7\) with \emph{non-overlapping} seed ranges over three seeds; \cref{sec:transfer-spec}), all at \(23.1\)M parameters versus \(90.5\text{--}90.6\)M for the ViT/DINOv2 decoders. \Cref{fig:pareto} shows the feature axis is Pareto-dominant on both segmentation and classification, while DINOv2 sits far to the right without a commensurate accuracy advantage under matched transfer.

\begin{table}[tb]
\centering
\caption{Segmentation transfer (\(256{\times}256\), matched U-Net decoder) on the \emph{held-out test} split. Dice/mIoU (\%) $\uparrow$; HD95 (px) $\downarrow$. \best{Bold}: best per task. $^{\dagger}$Near-saturated binary CT tasks (Lung, Colon) yield degenerate HD95 near zero for several models; see text. $^{\ddagger}$Prostate has a small test set (\(n{=}77\) slices) with high seed variance (std \(15\text{--}18\) Dice); we draw no task-level conclusion from it.}
\label{tab:seg}
\setlength{\tabcolsep}{4pt}\footnotesize
\resizebox{\linewidth}{!}{%
\begin{tabular}{llccc@{\hspace{6pt}}llccc}
\toprule
Dataset & Model & Dice & mIoU & HD95 & Dataset & Model & Dice & mIoU & HD95\\
\midrule
\multirow{3}{*}{BUSI} & ViT-UNet & 53.1 & 43.9 & 47.7 & \multirow{3}{*}{Lung} & ViT-UNet & 95.9 & 95.9$^{\dagger}$ & 0.1$^{\dagger}$\\
 & DINOv2-UNet & 53.7 & 43.2 & 50.2 & & DINOv2-UNet & 96.0 & 96.0$^{\dagger}$ & \best{0.0}$^{\dagger}$\\
 & \best{FAD-ViT} & \best{77.7} & \best{70.5} & \best{21.4} & & \best{FAD-ViT} & \best{96.3} & \best{96.2} & 0.0\\
\midrule
\multirow{3}{*}{Brain} & ViT-UNet & 80.2 & 76.7 & 3.3 & \multirow{3}{*}{Pancreas} & ViT-UNet & 82.9 & 81.2 & 2.3\\
 & DINOv2-UNet & 77.4 & 74.0 & 3.9 & & DINOv2-UNet & 81.5 & 80.0 & 2.4\\
 & \best{FAD-ViT} & \best{83.0} & \best{79.6} & \best{2.5} & & \best{FAD-ViT} & \best{86.1} & \best{83.9} & \best{1.8}\\
\midrule
\multirow{3}{*}{Heart} & ViT-UNet & 82.9 & 76.6 & 4.0 & \multirow{3}{*}{Hepatic} & ViT-UNet & 68.8 & 65.9 & 6.8\\
 & DINOv2-UNet & 77.8 & 71.1 & 4.3 & & DINOv2-UNet & 65.1 & 62.7 & 8.0\\
 & \best{FAD-ViT} & \best{85.6} & \best{80.0} & \best{3.3} & & \best{FAD-ViT} & \best{73.2} & \best{69.4} & \best{6.1}\\
\midrule
\multirow{3}{*}{Liver} & ViT-UNet & 91.5 & 90.3 & 1.5 & \multirow{3}{*}{Spleen} & ViT-UNet & 94.3 & 92.8 & 1.1\\
 & DINOv2-UNet & 90.3 & 89.0 & 2.0 & & DINOv2-UNet & 89.5 & 87.6 & 1.5\\
 & \best{FAD-ViT} & \best{92.9} & \best{91.8} & \best{1.4} & & \best{FAD-ViT} & \best{96.9} & \best{95.8} & \best{0.5}\\
\midrule
\multirow{3}{*}{Hippo.} & ViT-UNet & 86.5 & 81.5 & \best{5.4} & \multirow{3}{*}{Colon} & ViT-UNet & 88.6 & 88.5 & 0.2$^{\dagger}$\\
 & DINOv2-UNet & 86.4 & 81.4 & 5.7 & & \best{DINOv2-UNet} & \best{90.0} & \best{90.0}$^{\dagger}$ & \best{0.0}$^{\dagger}$\\
 & \best{FAD-ViT} & \best{87.1} & \best{82.2} & 5.5 & & FAD-ViT & 89.5 & 89.3 & 0.4\\
\midrule
\multirow{3}{*}{Prostate$^{\ddagger}$} & ViT-UNet & 53.5 & 47.8 & 10.3 & \multirow{3}{*}{\shortstack[l]{Mean\\(11 tasks)}} & ViT-UNet & 79.8 & 76.5 & --\\
 & DINOv2-UNet & 44.0 & 40.5 & 11.5 & & DINOv2-UNet & 77.4 & 74.1 & --\\
 & \best{FAD-ViT} & \best{58.8} & \best{53.5} & \best{9.3} & & \best{FAD-ViT} & \best{84.3} & \best{81.1} & --\\
\bottomrule
\end{tabular}%
}
\end{table}

Grouping the segmentation tasks by acquisition modality (\cref{tab:modality}) shows that the advantage is not uniform but tracks image content. The feature axis gains most on ultrasound (\(+24.6\) Dice on this split, \(+20.7\) over three seeds), a low-contrast, speckle- and texture-dominated modality, less on MRI (\(+2.8\), four tasks), and least on high-contrast structural CT (\(+2.1\), six tasks). The margin therefore grows precisely where discrimination depends on feature correlations rather than sharp geometry, consistent with the feature-axis hypothesis.

\begin{table}[tb]
\centering
\caption{Segmentation transfer by modality (mean Dice \%, matched U-Net decoder). The feature-axis gain over spatial-axis ViT is largest on texture-dominant, low-contrast ultrasound and smallest on high-contrast structural CT.}
\label{tab:modality}
\setlength{\tabcolsep}{6pt}\footnotesize
\begin{tabular}{lcccc}
\toprule
Modality (\#tasks) & ViT-UNet & DINOv2-UNet & FAD-ViT & \(\Delta\) vs ViT\\
\midrule
Ultrasound (1) & 53.1 & 53.7 & \best{77.7} & \best{+24.6}\\
MRI (4)        & 75.8 & 71.4 & \best{78.6} & +2.8\\
CT (6)         & 87.0 & 85.4 & \best{89.1} & +2.1\\
\bottomrule
\end{tabular}
\end{table}

\begin{figure}[tb]
\centering
\includegraphics[width=1\linewidth]{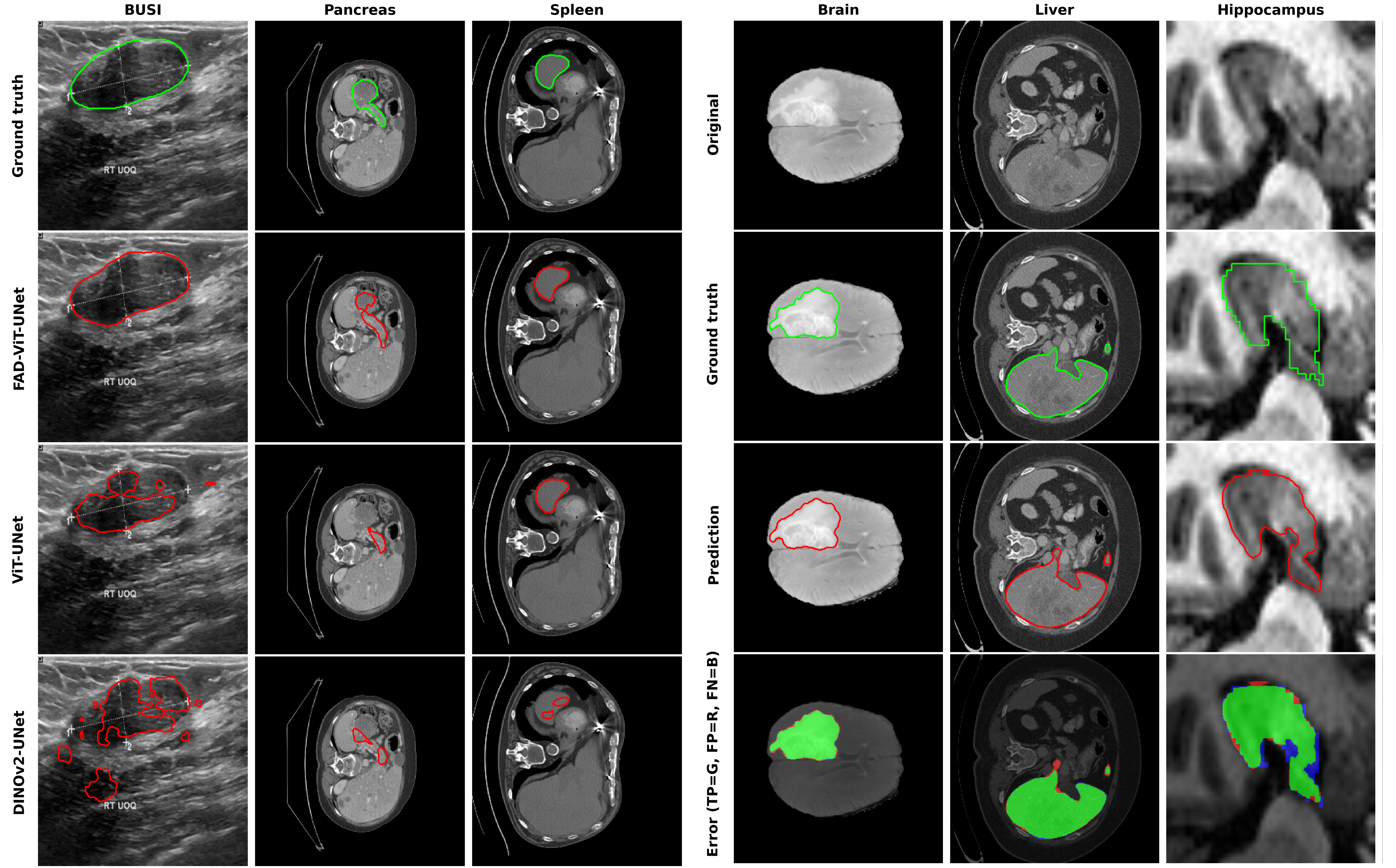}
\caption{Qualitative segmentation transfer. Feature-axis FAD-ViT-UNet produces cleaner boundaries and fewer spurious regions than spatial-attention baselines; error maps (TP green, FP red, FN blue) concentrate residual error at ambiguous boundaries.}
\label{fig:seg}
\end{figure}

\subsection{Channel-Primary Decoupling versus Coupled Channel Attention}
\label{sec:decouple}
Channel attention alone does not explain the transfer benefit; making it \emph{primary} and strictly decoupled does. We transfer the two closest channel-aware designs, XCiT~\cite{elnouby_xcit_2021} and DaViT~\cite{ding_davit_2022}, under the identical U-Net protocol. In both the axes stay entangled (pointwise \(Q/K/V\) pre-mix channels; spatial mixing persists), so pretraining cannot be attributed to one axis; FAD-ViT, with depthwise-only \(Q/K/V\) and no auxiliary spatial modules, is the minimal design that removes this confound. All three channel-aware backbones beat the pure spatial-axis ViT-UNet on the held-out ultrasound test by \(+22\) to \(+25\) Dice (\cref{tab:decouple}; extended to all eleven tasks in the supplement), so feature-axis interaction drives the gain. The three are close (within \(2.6\) Dice) yet invert the natural-image ordering: XCiT and DaViT match or beat FAD-ViT on ADE20K but only tie it under medical transfer, where FAD-ViT reaches the strongest coupled design's accuracy (\(77.7\) vs.\ \(77.8\) Dice) at the fewest parameters (\(23.1\) vs.\ \(27.1\) and \(34.4\)M). That FAD-ViT matches rather than beats these coupled designs supports our thesis: the benefit comes from the shared feature axis, not from any one architecture. FAD-ViT is simply its efficient, attributable endpoint, transferring at least as well as heavier coupled designs without their machinery.

\begin{table}[tb]
\centering
\caption{Channel-primary decoupling versus coupled channel attention, matched U-Net decoder. Natural-image competence (ADE20K mIoU) versus medical transfer (held-out BUSI test Dice \%). Coupled backbones (XCiT, DaViT) match or beat FAD-ViT on natural images but only tie it on medical transfer, which the decoupled design reaches at the lowest parameter count. \best{Bold}: best.}
\label{tab:decouple}
\setlength{\tabcolsep}{5pt}\footnotesize
\begin{tabular}{llccc}
\toprule
\multirow{2}{*}{Backbone} & \multirow{2}{*}{Global operator} & Params & Natural & Medical\\
 & & (M) & mIoU & BUSI Dice\\
\midrule
ViT-UNet     & spatial-token only            & 90.6 & 41.9 & 53.1\\
XCiT-UNet    & channel + spatial (columnar)  & 27.1 & \best{43.9} & 75.1\\
DaViT-UNet   & channel/spatial interleaved   & 34.4 & 42.8 & \best{77.8}\\
\rowcolor{oursgray}
FAD-ViT-UNet & feature-axis only (decoupled) & \best{23.1} & 42.6 & 77.7\\
\bottomrule
\end{tabular}
\end{table}

\subsection{Low-Data Linear Probing}
\label{sec:lowdata}
Freezing the backbone and training only a linear head isolates the linear separability of the pretrained features at 1\%, 10\%, and 100\% labels (\cref{tab:probe}). The result is dataset-dependent: the feature axis leads on PathMNIST, OCTMNIST, PneumoniaMNIST, and BloodMNIST, with the margin largest in the 1\% regime where separability matters most, while DINOv2 and ViT-S/16 lead on a few others. Feature-axis pretraining thus yields more linearly separable medical features under scarce supervision on texture- and intensity-dominated targets, though not uniformly.

\section{Analysis: Isolating the Attention Axis}
\label{sec:analysis}

\subsection{Ablations}
\Cref{tab:ablation} ablates the two block-defining choices from scratch, removing pretraining and capacity as confounds (the feature- and spatial-axis backbones have \emph{identical} \(23.1\)M parameters; disabling the spatial tail gives \(16.4\)M). From random initialization the axis swap costs only \(1.1\text{--}1.8\) Dice and removing the spatial tail \(1.1\text{--}1.3\) Dice, within from-scratch seed variance: absent pretraining, the feature axis is at most a marginal architectural advantage. This near-parity is the control that gives the transfer result its force, since the same axis that is neutral from scratch becomes decisive once pretrained (\cref{sec:transfer-spec}). The axis does not make the network intrinsically stronger; it makes the pretrained representation more transferable. We hold the operator parameterization fixed (single-head channel-token attention as in DaViT~\cite{ding_davit_2022}/SHViT~\cite{yun_shvit_2024}; depthwise \(Q/K/V\) per the decoupling requirement, \cref{sec:method}) so the axis is the only variable; BUSI and Heart bracket the two modality extremes (\cref{tab:modality}).

\begin{table}[tb]
\centering
\caption{Low-data linear probing (frozen backbone). Test Accuracy (\%) at 1/10/100\% labels (single-seed; 3-seed means in Table~S9). \best{Bold}: best per row and fraction.}
\label{tab:probe}
\setlength{\tabcolsep}{4pt}\footnotesize
\resizebox{\linewidth}{!}{%
\begin{tabular}{lccccccccc}
\toprule
\multirow{2}{*}{Dataset} & \multicolumn{3}{c}{FAD-ViT (feat-axis)} & \multicolumn{3}{c}{ViT-S/16} & \multicolumn{3}{c}{DINOv2-B}\\
\cmidrule(lr){2-4}\cmidrule(lr){5-7}\cmidrule(lr){8-10}
 & 1\% & 10\% & 100\% & 1\% & 10\% & 100\% & 1\% & 10\% & 100\%\\
\midrule
PathMNIST      & 79.8 & \best{85.3} & \best{86.2} & 74.6 & 82.3 & 84.5 & \best{80.0} & 70.5 & 80.3\\
DermaMNIST     & 66.7 & \best{69.2} & 68.8 & 66.7 & \best{69.2} & \best{70.1} & \best{66.9} & 68.8 & \best{70.1}\\
OCTMNIST       & \best{73.0} & \best{76.3} & \best{76.5} & 71.2 & 75.3 & 75.8 & 68.3 & 74.7 & 75.7\\
PneumoniaMNIST & \best{87.1} & 88.7 & 89.1 & 71.2 & 81.0 & 80.5 & 74.4 & \best{89.4} & \best{89.6}\\
RetinaMNIST    & 44.2 & 46.7 & 44.6 & \best{48.8} & \best{50.8} & \best{52.5} & 42.1 & 43.3 & 46.7\\
BreastMNIST    & 71.8 & 70.9 & \best{73.5} & \best{73.5} & \best{72.6} & 70.1 & 67.5 & 71.8 & 71.8\\
BloodMNIST     & 51.1 & \best{75.7} & \best{79.3} & 48.2 & 71.1 & 73.4 & \best{51.4} & 70.7 & 74.8\\
OrganAMNIST    & 72.0 & 84.0 & 85.6 & 62.9 & 80.7 & 82.4 & \best{72.9} & \best{86.7} & \best{88.2}\\
\bottomrule
\end{tabular}%
}
\end{table}

\begin{table}[tb]
\centering
\caption{Medical ablation on BUSI and Heart, from-scratch Dice (\%), matched decoder and schedule. \(\Delta\) is relative to the full model on the same task. From scratch the axis swap and the spatial tail each cost only \(1\text{--}2\) Dice at identical or lower capacity; the decisive axis effect appears under transfer (\cref{tab:transfer-gap}), not here. \best{Bold}: full model.}
\label{tab:ablation}
\setlength{\tabcolsep}{5pt}\footnotesize
\resizebox{\linewidth}{!}{%
\begin{tabular}{lcccc}
\toprule
Variant & BUSI Dice & \(\Delta\) & Heart Dice & \(\Delta\)\\
\midrule
\best{Full FAD-ViT (feat-axis, dw-\(Q/K/V\), +tail)} & \best{64.2} & -- & \best{82.6} & --\\
\;$\rightarrow$ swap to spatial-axis attn.\ (23.1M) & 63.1 & \(-1.1\) & 80.8 & \(-1.8\)\\
\;$\rightarrow$ remove spatial tail (16.4M) & 63.1 & \(-1.1\) & 81.3 & \(-1.3\)\\
\bottomrule
\end{tabular}}
\end{table}

\subsection{Transfer Specificity}
\label{sec:transfer-spec}
A higher downstream score alone cannot separate a better architecture from a better \emph{transfer}, nor either from raw capacity. We disentangle these by training backbones from both random initialization and ImageNet-1K pretraining and reading off the transfer gap, pretrained minus from-scratch Dice, on breast ultrasound (\cref{tab:transfer-gap}). To control the natural confound we match a spatial-axis ViT to FAD-ViT in parameters (ViT-S/16-UNet, \(22.9\) vs.\ \(23.1\)M): over three seeds it converts pretraining into only \(+4.3\) Dice against \(+15.1\) for the feature axis, roughly a threefold difference. Scaling the spatial baseline to ViT-B (\(90.6\)M) or using a pretrained convolutional encoder (ResNet50-UNet, \(71.7\)M) still yields far smaller gaps (\(+5.7\), \(+6.0\)). At or below the capacity of every baseline the feature axis recovers two-to-threefold more of the pretrained representation on this target: the operator, not the parameter budget, governs what survives. Source-task quality does not account for this either: the backbones' ImageNet-1K top-1 accuracies are comparable (FAD-ViT \(81.6\), ViT-S/16 \(78.8\), ViT-B \(79.2\)), the small edge favouring the feature axis by only \({\approx}2.5\) points, far too little to produce \(10\text{--}13\) Dice transfer differences.

\begin{table}[tb]
\centering
\caption{Transfer-specificity on breast ultrasound (held-out test Dice \%): from scratch vs.\ ImageNet-1K pretraining, matched U-Net decoder and schedule. Gap $=$ pretrained $-$ scratch. Feature-axis, ViT-S, and ViT-B values are averaged over seeds (ViT-S over three); ResNet50 is single-seed.}
\label{tab:transfer-gap}
\setlength{\tabcolsep}{5pt}\footnotesize
\begin{tabular}{llccc}
\toprule
Backbone & Params (M) & Scratch & Pretrained & Gap\\
\midrule
\rowcolor{oursgray}
Feature-axis FAD-ViT      & 23.1 & 61.3 & \best{76.4} & \best{+15.1}\\
Spatial ViT-S (matched)   & 22.9 & 48.9 & 53.2 & +4.3\\
Spatial ViT-B             & 90.6 & 49.9 & 55.7 & +5.7\\
Convolutional ResNet50    & 71.7 & 61.7 & 67.7 & +6.0\\
\bottomrule
\end{tabular}
\end{table}

A single dataset cannot establish that the gap is a property of the operator rather than of BUSI. With the strict in-place axis swap (identical \(23.1\)M skeleton, only the global-mixing axis flipped) we measure the same gap on one representative task per modality (\cref{tab:crossmod}): the feature axis converts ImageNet pretraining into a far larger gain than the spatial axis on texture-dominated ultrasound and retains a smaller but consistent edge on abdominal CT and cardiac MRI, also attaining higher absolute test Dice at matched capacity. The advantage thus tracks the modality trend of \cref{tab:modality}: largest where appearance and texture discriminate, shrinking without reversing where sharp geometry already fixes the boundary (matched-capacity results for all eleven tasks, and the completed axis--tail ablation, are in the supplement).

\begin{table}[tb]
\centering
\caption{Cross-modality transfer gap under the strict in-place axis swap (identical \(23.1\)M skeleton, only the global-mixing axis flipped). Held-out test Dice (\%); Gap $=$ pretrained $-$ scratch. CT/MRI are multi-seed means (std \(\le\!2.8\)); ultrasound reproduces the primary run (\cref{tab:transfer-gap}). The feature axis wins the gap on every modality, most on texture-dominated ultrasound.}
\label{tab:crossmod}
\setlength{\tabcolsep}{5pt}\footnotesize
\begin{tabular}{lcccccc}
\toprule
\multirow{2}{*}{Modality (task)} & \multicolumn{3}{c}{Feature axis (ours)} & \multicolumn{3}{c}{Spatial axis}\\
\cmidrule(lr){2-4}\cmidrule(lr){5-7}
 & Scratch & Pretr. & Gap & Scratch & Pretr. & Gap\\
\midrule
\rowcolor{oursgray}
Ultrasound (BUSI) & 61.3 & \best{76.4} & \best{+15.1} & 63.1 & 65.3 & +2.2\\
CT (Spleen)       & 90.6 & \best{95.3} & \best{+4.7} & 91.1 & 94.3 & +3.2\\
MRI (Heart)       & 80.9 & \best{84.5} & \best{+3.6} & 80.5 & 82.2 & +1.7\\
\bottomrule
\end{tabular}
\end{table}

\subsection{Why Feature-Axis Attention Transfers}
\label{sec:why}
\begin{figure}[tb]
\centering
\includegraphics[width=1\textwidth]{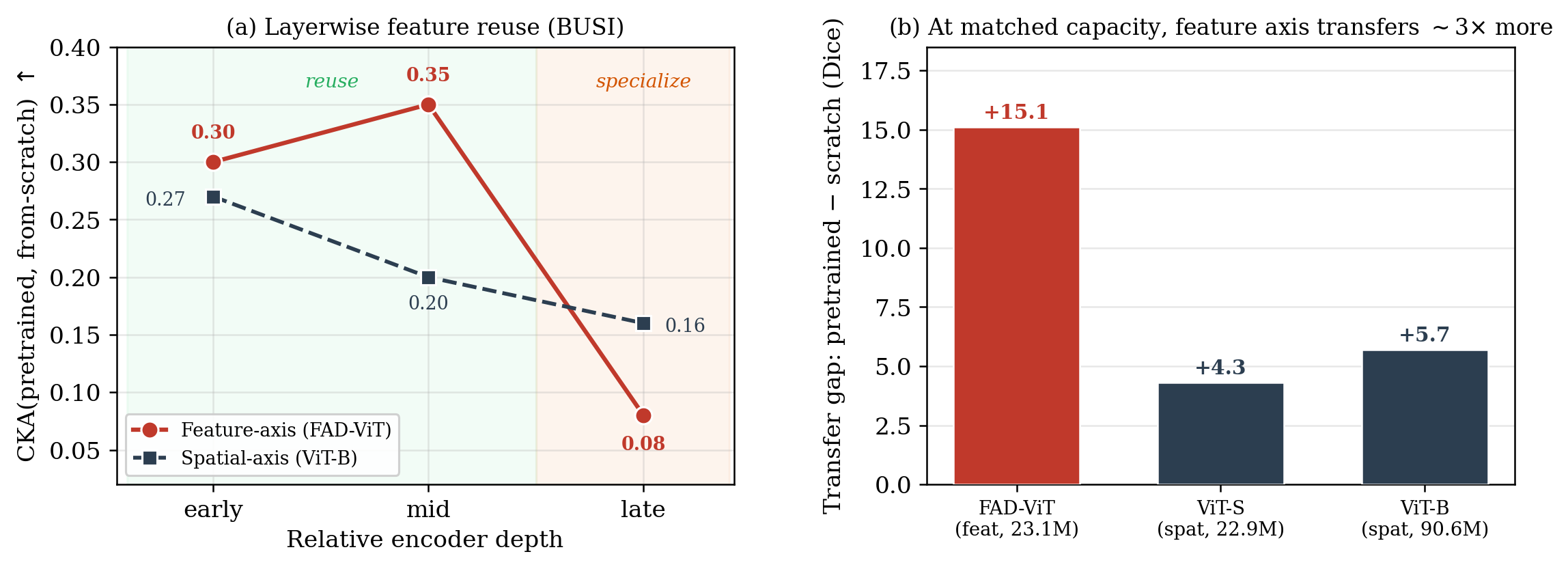}
\caption{Feature reuse tracks the transfer gap. (a) Linear CKA between the ImageNet-pretrained and from-scratch encoders on BUSI (higher = more reuse): the feature axis reuses more early/mid structure and specializes hardest late, whereas the spatial axis decays monotonically. (b) Their transfer gap (pretrained \(-\) scratch Dice): higher mid-level reuse coincides with a larger gain from identical pretraining.}
\label{fig:cka}
\end{figure}

The mechanism follows from what each axis pretrains. Feature-axis attention mixes channels globally, so pretraining shapes channel-correlation statistics, texture, intensity, and co-activation, that recur across scanners and anatomy; spatial-token attention instead shapes object-layout priors specific to natural scenes and largely invalidated by medical domain shift. Centered kernel alignment (CKA)~\cite{kornblith_cka_2019} between the pretrained and from-scratch encoders on BUSI is consistent with this (\cref{fig:cka}): feature-axis alignment stays higher at early and mid depth and collapses at the last stage, carrying over general channel structure while relearning only the task-specific top, whereas spatial-axis alignment decays monotonically, the profile of a weaker transfer prior. We read CKA as a corroborative correlate of the measured transfer gap, not direct proof of feature reuse. The same axis yields cleaner segmentation boundaries and fewer spurious components on texture-dominant targets (\cref{fig:seg}).

\section{Conclusion and Future Work}
We treated the axis of global attention as an experimental variable and introduced a strictly axis-decoupled backbone in which depthwise operators route all spatial structure and channel-token attention performs all global mixing. Moving global interaction from spatial to feature tokens converts identical ImageNet pretraining into several times larger downstream gains at matched capacity, outperforming spatial backbones across diverse segmentation tasks at a quarter of the parameters while reusing more pretrained early and mid-level structure, the signature of a transfer prior rather than a capacity effect. The gain is largest on low-contrast ultrasound and smaller but consistent on MRI and CT. Feature-centric global attention is thus a cheap, principled design choice for medical backbones; natural next steps are stronger pipelines (nnU-Net), other architecture families, and self-supervised pretraining.

\subsubsection*{Acknowledgements.}
This publication has emanated from research conducted with the financial support of Taighde \'Eireann -- Research Ireland under Grant number 18/CRT/6183.

\clearpage
\bibliographystyle{splncs04}
\bibliography{references}

\end{document}